\documentclass[conference]{IEEEtran}
\IEEEoverridecommandlockouts
\usepackage{cite}
\usepackage{amsmath,amssymb,amsfonts}
\usepackage{algorithmic}
\usepackage{graphicx}
\usepackage{textcomp}
\usepackage{xcolor}
\def\BibTeX{{\rm B\kern-.05em{\sc i\kern-.025em b}\kern-.08em
    T\kern-.1667em\lower.7ex\hbox{E}\kern-.125emX}}

\usepackage{makecell}
\usepackage{subcaption}
\usepackage{etoolbox}
\renewcommand{\bfseries}{\fontseries{b}\selectfont}
\robustify\bfseries
\newrobustcmd{\B}{\bfseries}

\usepackage[acronym, style=super, section=section, nonumberlist, nomain, nopostdot]{glossaries}
\usepackage[]{glossaries-extra}
\usepackage{url}
\usepackage{multirow}
\usepackage{arydshln}
\usepackage{tikz}
\usepackage{fancyhdr}
\usepackage{hyperref}

\newcommand\copyrighttext{%
  \footnotesize © 2024 IEEE.  Personal use of this material is permitted.  Permission from IEEE must be obtained for all other uses, in any current or future media, including reprinting/republishing this material for advertising or promotional purposes, creating new collective works, for resale or redistribution to servers or lists, or reuse of any copyrighted component of this work in other works.
}

\newcommand\copyrightnotice{%
  \begin{tikzpicture}[remember picture,overlay]
    \node[anchor=south,yshift=10pt] at (current page.south) 
      {\fbox{\parbox{\dimexpr\textwidth-\fboxsep-\fboxrule\relax}{\copyrighttext}}};
  \end{tikzpicture}%
}

\newsavebox{\largestimage}

\fancypagestyle{firstpage}{
  \fancyhf{} 
  \fancyhead[L]{%
    \footnotesize
    S. Krylova, F. Schmidt and V. Vlassov, "Leveraging Machine Learning Models to Predict the Outcome of Digital Medical Triage Interviews,"\\
    \textit{2024 International Conference on Machine Learning and Applications (ICMLA)}, Miami, FL, USA, 2024, pp. 160–167,\\
    doi: \href{https://doi.org/10.1109/ICMLA61862.2024.00028}{10.1109/ICMLA61862.2024.00028}
  }
}

\newacronym{AI}{AI}{Artificial Intelligence}
\newacronym{ML}{ML}{Machine Learning}
\newacronym{DL}{DL}{Deep Learning}
\newacronym{RGS}{RGS}{Rådgivningsstödet}
\newacronym{EU}{EU}{European Union}
\newacronym{MDR}{MDR}{Medical Device Regulation}
\newacronym{GSPR}{GSPR}{General Safety and Performance Requirements}
\newacronym{HPO}{HPO}{Hyperparameter optimisation}
\newacronym{CSV}{CSV}{Comma-separated values}
\newacronym{MLP}{MLP}{Multilayer perceptron}
\newacronym{GBDT}{GBDT}{Gradient-boosted decision trees}
\begin{document}

\title{Leveraging Machine Learning Models to Predict the Outcome of Digital Medical Triage Interviews
}
\author{\IEEEauthorblockN{Sofia Krylova\IEEEauthorrefmark{1}\IEEEauthorrefmark{2},
Fabian Schmidt\IEEEauthorrefmark{2},
Vladimir Vlassov\IEEEauthorrefmark{2}}
\IEEEauthorblockA{\IEEEauthorrefmark{1}Platform24 AB, Stockholm, Sweden}
\IEEEauthorblockA{\IEEEauthorrefmark{2}School of Electrical Engineering and Computer Science,
KTH Royal Institute of Technology, Stockholm, Sweden}
Email: \IEEEauthorrefmark{1}sofia.krylova@platform24.com, \IEEEauthorrefmark{2}\{krylova, fschm, vladv\}@kth.se 
}

\maketitle
\thispagestyle{firstpage}
\copyrightnotice
\pagestyle{plain}
\begin{abstract}
One of the key advances in digital healthcare is the implementation of digital triage, which, using online tools, web, and mobile apps, allows for efficient assessment of patient needs, prioritizing cases, and directing them to appropriate healthcare services. 
Many existing digital triage systems are questionnaire-based, guiding patients to appropriate care levels based on information (e.g., symptoms, medical history, and urgency) provided by the patients answering questionnaires. Such a system often uses a deterministic model with predefined rules to determine care levels. It faces challenges with incomplete triage interviews since it can only assist patients who finish the process. In this study, we explore the use of machine learning (ML) to predict outcomes of unfinished interviews, aiming to enhance patient care and service quality. Predicting triage outcomes from incomplete data is crucial for patient safety and healthcare efficiency. 




Our findings show that decision-tree models, particularly LGBMClassifier and CatBoostClassifier, achieve over 80\% accuracy in predicting outcomes from complete interviews while having a linear correlation between the prediction accuracy and interview completeness degree. For example, LGBMClassifier achieves 88,2\% prediction accuracy for interviews with 100\% completeness, 79,6\% accuracy for interviews with 80\% completeness, 58,9\% accuracy for 60\% completeness, and 45,7\% accuracy for 40\% completeness. The TabTransformer model demonstrated exceptional accuracy of over 80\% for all degrees of completeness but required extensive training time, indicating a need for more powerful computational resources. The study highlights the linear correlation between interview completeness and predictive power of the decision-tree models.

\end{abstract}


\section{Introduction}

With the rise of digital health solutions, such as smartphone health and medical apps, health wearables, and telehealth platforms, including digital triage, healthcare delivery has been significantly transformed to enhance health and well-being. 

Medical triage is the initial point of contact within the healthcare system, handling medical conditions for children and adults. Triage allows assessing patient conditions and needs, prioritizes cases, and directs patients to appropriate healthcare services. Emergency departments and primary care employ triage, but the specifics and goals of triage differ. Patients in primary care usually do not require urgent help, so this triage system focuses more on efficiently utilizing healthcare resources \cite{stiernstedt_digifysiskt_2019}. In this work, we focus on primary care triage. Common challenges related to manual triage include a need for clinical competency and psychological capabilities and challenges in human resources management and performance \cite{bijani_challenges_2019}.
Digital triage uses online tools, web, and mobile apps to efficiently collect patients' information (e.g., symptoms, medical history, urgency), assess their conditions, and guide them to appropriate care levels. Digital triage addresses resource shortages, improves patient care, and enhances service accessibility and efficiency. 


For example, Platform24 company developed Triage24, a questionnaire-based digital triage system, to tackle the abovementioned challenges. This online self-service tool directs patients to the appropriate level of care, time, and location based on their healthcare needs, urgency, and other information acquired through a thoroughly designed questionnaire.
Platform24 uses a deterministic model for primary care triage to define a level of care based on patient answers to a corresponding questionnaire. However, this approach can only serve finished interviews. 

This paper investigated the potential of machine learning (ML) models to predict the outcomes of unfinished triage interviews, aiming to enhance patient care and service quality.

We formulate the question-based digital triage as a classification task on sparse data (triage interviews) and evaluate various ML and ML models, like decision trees and attention-based models, to assess their applicability and identify those models that provide the highest accuracy in predicting the outcomes of incomplete triage interviews. The scope of the work included data anonymization (not considered here) and transformation into a data schema suitable for ML purposes, testing models like LGBMClassifier (implemented within the LightGBM framework \cite{10.5555/3294996.3295074}), CatBoostClassifier (implemented within the CatBoost framework \cite{dorogush2018catboost}), TabTransformer, and other classifiers, shortlisting the best-performing models, and evaluating them on unfinished interviews. We also researched the correlation between ML models' completeness degree and accuracy. Thus, the main research question of this work is: Which ML and DL model(s) can effectively predict a level of care based on patients' unfinished interviews to provide a patient with a proper level of care?

\begin{figure*}[!ht]
  \begin{center}
    \includegraphics[width=0.86\textwidth]{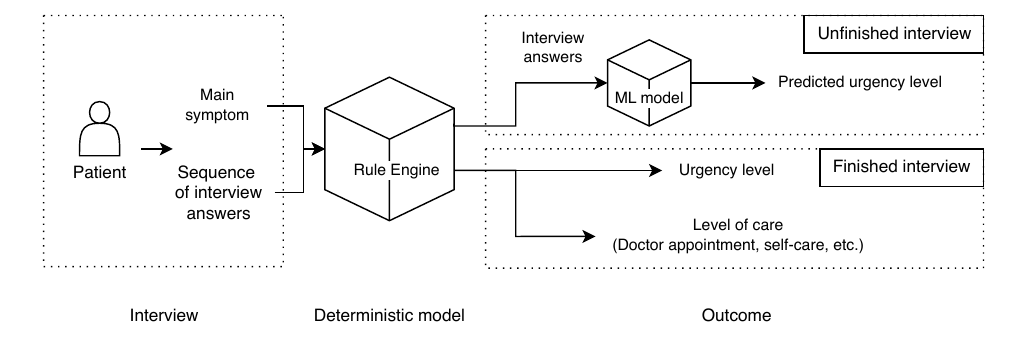}
  \end{center}
  \caption{System integration of the ML model into the existing triage}
  \label{fig:integration}
\end{figure*}

Leveraging the strengths of predictive ML models can enhance the efficiency and accuracy of digital triage, ultimately improving patient outcomes and healthcare delivery. We envision combining a predictive ML model with a deterministic model in a questionnaire-based digital triage system as a side-car application, as depicted in Figure~\ref{fig:integration}. The ML system can be used as a sub-workflow triggered only when a patient tries to exit triage before finishing an interview, i.e., in the case of an unfinished, incomplete interview. This way, the rule engine of the deterministic model can send 
answers from an incomplete interview to the 
ML model that can predict the triage outcome and present it to the patient, who can decide whether to continue an interview and get the final outcome or exit without a triage result. 

In this work, we consider Triage24 by Platform24 as a case study to investigate which ML models can accurately predict the outcomes of unfinished interviews to enhance a questionnaire-based triage like Triage24. Using a real-world dataset of triage interviews from Platform24, we evaluate various ML and DL models, including decision trees and TabTransformer, to identify those with the highest accuracy in predicting incomplete triage outcomes. 

Being catered to a specific patient need, individual interviews possess extremely diverse features, which makes such datasets highly sparse. By exploring various models capable of handling sparse data, such as LGBMClassifier and CatBoostClassifier, the study seeks to compare different solutions that could improve the efficiency of remote triage systems.

We train the 
selected ML and DL models on a dataset of complete interviews with ground-truth outcomes designed by a team of medical content creators and certified under MDR (Medical Device Regulation). As a triage interview is a set of question-answer pairs, an incomplete interview is a sub-set of generally several complete interviews with different outcomes, which makes it difficult, if not possible at all, to define a ground-truth outcome for the incomplete interview. Constructing a dataset of incomplete interviews and using the original outcomes for training would be speculative rather than based on ground truth, inevitably introducing inaccuracy and decreasing the model's reliability. 







In summary, this paper makes the following contributions:
\begin{itemize}
    \item We have constructed a training tabular dataset of completed triage interviews from a real-world raw dataset of completed interviews with outcomes defined by the deterministic model of an operational questionnaire-based triage system. This work included data anonymization (not considered here) and cleaning, transforming interview records into a data schema suitable for ML purposes, and data splitting.

    \item We have selected and trained several commonly used ML models, including five tree-based ML models and a transformer model, on a sparse, highly dimensional training dataset of completed triage interviews.
    
    \item We have performed a systematic comparison of the selected ML models in terms of their performance in terms of training and test accuracy, as well as training and inference times on datasets.
    \item We have constructed a dataset of unfinished interviews with various degrees of completeness.
    
    \item We have assessed the top-performing ML models on the dataset of unfinished interviews in terms of inference accuracy and analyzed the relationship between the degree of completeness and the predictive accuracy of the models.
\end{itemize}

The remainder of this paper is structured as follows. Section~\ref{sec:background} presents relevant information about triage systems and describes related work on digital solutions in triage and state-of-the-art models tailored to handle sparse data. Section~\ref{sec:methods} presents the methodology and method used to solve the problem and evaluate the solution. Section~\ref{sec:results} presents and discusses the results of each experimental round and the final evaluation. Section~\ref{sec:discussions} discusses the study results and limitations. Section~\ref{sec:conclusion} concludes that paper and points out directions for future work.

\section{Background and related work}
\label{sec:background}



Online triage typically involves structured questionnaires to comprehensively analyze the patient's conditions and symptoms. The process consists of going through questions and helping the patient describe his symptoms, which are then forwarded to healthcare professionals for review during working hours. If needed, the healthcare practitioner may inquire for further information to direct the patient to the appropriate specialist \cite{eldh_health_2020}. This process takes time and requires the availability of healthcare practitioners \cite{kang_machine_2021}. Recently, there has been a push to utilize digital tools to expedite triaging \cite{miles2020using} similar to recent work in predicting a diagnosis based on test results and anamnesis \cite{miotto_deep_2018,wang_disease-prediction_2023,rasmy_med-bert_2021,kang_machine_2021,jones_artificial_2022,jammeh_machine-learning_2018}.

With the impact of rapid technological development in recent years, many healthcare providers have started looking into possible ways of implementing modern solutions, namely \gls{AI} and \gls{ML}. A few research studies investigated the performance of \gls{AI} in a healthcare setting and showed the potential for improving patient experience \cite{adebayo_exploring_2023,gkouskos_exploring_2023,state_designing_2020,SANCHEZSALMERON_2022_ML_Triage,defilippo2024leveraging}. As a result of continuous healthcare digitalization, new digital triage systems have emerged that allow for increased efficiency in healthcare services by delegating the triage task from a healthcare professional to the patient.

Platform24's Triage24 is a notable example of guiding patients to appropriate care levels through a questionnaire-based system.
Widely used across Sweden, Triage24 successfully serves patients who have completed their triage process but face the challenge of defining the outcome of incomplete interviews. Overcoming this limitation would enable the system to support these patients and enhance the service quality. Hence, a solution is needed to predict the level of care for unfinished interviews.

\section{Method}
\label{sec:methods}
\subsection{Data collection and preparation}

\label{sec:dataPreparation}



This study uses the real-world dataset of about 330,000 complete triage interviews with definite outcomes conducted in 2023 in the Triage24 system. Each interview comprises, on average, 12 questions with corresponding answers, which means that the dataset contains, in total, approximately 4 million answers. 

\begin{figure*}
  \begin{center}
    \includegraphics[width=.75\textwidth]{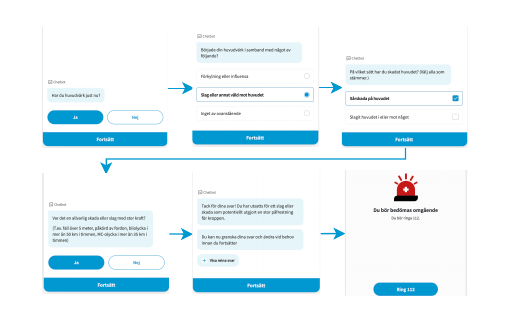}
  \end{center}
  \caption{An example of a Triage24 interview (in Swedish) with the \texttt{immediate} urgency outcome. 
  English: 1. Do you have a headache right now?: \textbf{Yes}/No. 2. Did your headache start in connection with any of the following?: Cold or flu/\textbf{Blows or other violence to the head}/None of the above. 3. In what way have you injured your head? (Choose all that apply.): \textbf{Injury to the head}/Hit your head on or against something. 4. Was it a serious injury or impact with great force? (E.g., fall over 5 meters, hit by a car, car accident at more than 50 km/h, motorcycle accident at more than 35 km/h): \textbf{Yes}/No. 5. Thank you for your answers! You have suffered a blow or injury that has potentially put a lot of strain on your body. You can now review your answers and others if necessary before continuing. 6. You should be assessed immediately. You should call 112.}
  \label{fig:triageimmediate}
\end{figure*}

\begin{figure*}
  \begin{center}
  \savebox{\largestimage}{\includegraphics[height=.35\textheight]{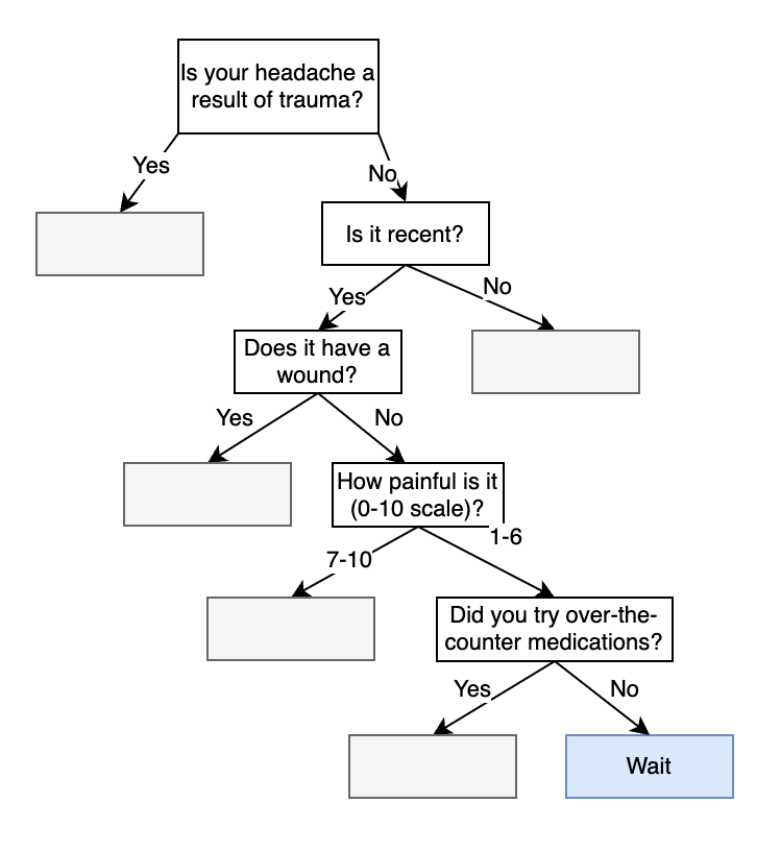}}
    \begin{subfigure}{0.49\textwidth}
        \begin{center}
        \usebox{\largestimage}
        \end{center}
        \caption{Patient 1 interview with an \texttt{wait} outcome}
        \label{fig:p1}
    \end{subfigure}
\hfill
    \begin{subfigure}{0.49\textwidth}
        \begin{center}
    \raisebox{\dimexpr.5\ht\largestimage-.5\height}{\includegraphics[width=\textwidth]{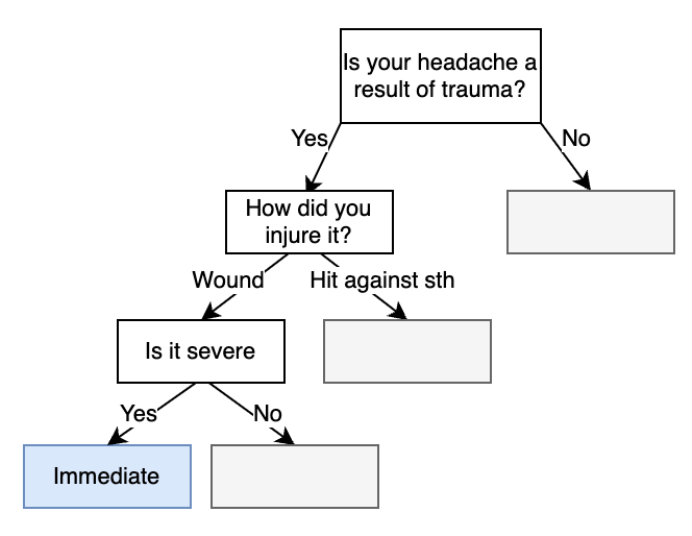}}
        \end{center}
        \caption{Patient 2 interview with a \texttt{immediate} outcome}
        \label{fig:p2}
    \end{subfigure}
    \end{center}
    \caption{Two patient interviews with different outcomes based on the questionnaires}
    \label{fig:patients}
\end{figure*}

Each interview comprises questions of various types, e.g., True/False, numeric, textual, single, or multiple choice, from different questionnaires based on the patient's situation. A questionnaire is a set of questions about a symptom. For example, a headache questionnaire contains questions on the cause of the headache, such as trauma, severity level, etc. Figure~\ref{fig:triageimmediate} shows the process of a patient answering a questionnaire. Since every interview instance is customized to the patient's answers, the sequence of questions is dynamically constructed for each interview based on previous answers. Figure~\ref{fig:patients} shows examples of two different interview instances for the same main symptom - headache. Depending on the answers, an interview instance can also lead to questions about other symptoms, e.g., breathing problems. Such a tailored approach to patient care leads to high interview variability. 

As a result, the datasets present high sparsity: the full dataset has over 4,000 columns, and in every row, only 10-20 columns contain non-null values. We removed all empty and duplicated records were removed. Text questions were also excluded since they are not used to define the outcome in the original deterministic model but rather used to provide additional information to a healthcare practitioner.
As a result, the datasets present high sparsity: the full dataset has over 4,000 columns, and in every row, only 10-20 columns contain non-null values. We removed all empty and duplicated records. We excluded text questions since they are not used to define the outcome in the original deterministic model but to provide additional information to a healthcare practitioner.

The data sampling step included creating three main datasets:
\begin{enumerate}
    \item Experimental dataset: a small subset (approximately 15,000 interviews and over 2,000 features) for model selection and testing.
    \item Training dataset: a dataset that has 80\% of collected data (approximately 265,000 interviews and over 4,000 features) used for training and validating selected models.
    \item Evaluation dataset: a 20\% dataset used to construct a set of unfinished interviews with various degrees of completeness for the final evaluation of the models' performance.
\end{enumerate}

Since the source data contains highly sensitive information, we anonymized it at the collection step. The dataset included only interview records of adult ($>$18 years old) patients - we omitted children from the study. We carefully checked the dataset to ensure it did not contain any personal information (name, personal number, age, etc.). New dummy IDs replace interview and answer IDs to eliminate any opportunity to track down an interview and retrieve patient data.

\subsection{Data analysis and feature engineering}
The target class of this study is an interview outcome. The target class represents the urgency level of the following classes: \texttt{wait}, \texttt{planned}, \texttt{promptly}, \texttt{immediate}, and \texttt{acute}. We subsequently converted these categories into a numerical representation. The dataset exhibits a slight imbalance, which we addressed by employing suitable metrics during the evaluation phase.

Due to the sparse nature of the dataset, we apply two approaches in this study. The first approach involved leaving empty (NaN) values and utilizing models that can handle the sparsity. NaN values may carry additional information relevant to the domain. This dataset is referred to as the \textit{NaN dataset} throughout the study. The second approach involved replacing empty values with 0/False and using models that work only with dense datasets. However, this method has the drawback of introducing additional noise. We refer to this dataset as the \textit{Non-NaN dataset} in this study. Table~\ref{tab:transformednonnanstructure} shows an example without the categorical transformation.

\begin{table}
  \begin{center}
    \resizebox{\columnwidth}{!}{%
    \begin{tabular}{cccccc}
      \makecell{\B Dummy \\ \B Interview ID} & \textbf{Outcome} & \makecell{ \B injury-triage-\\ \B pain-level} & \makecell{\B injury-triage-\\ \B pain-location} & \textbf{ ... } & \makecell{\B headache-triage-\\ \B counter-meds} \\
      \hline
      1 & promptly & 5 & shoulder & ... & False \\
      \hline
      2 & wait & 0 & none & ... & False \\
    \end{tabular}
    }
    \caption{Transformed non-NaN dataset structure.}
    \label{tab:transformednonnanstructure}
  \end{center}
\end{table}

\subsection{ML Model Selection and Justification}
Sparse, questionnaire-based data often results in high-dimensional feature spaces, where the number of features far exceeds the number of samples. This high dimensionality can pose challenges for model selection as it increases the risk of overfitting, where models learn noise or irrelevant patterns from the data. Traditional ML models may struggle to generalize effectively or require specialized preprocessing techniques to handle missing values, such as imputation or feature engineering. Furthermore, questionnaire-based data often captures complex relationships and interactions between variables, which may not be linear or easily interpretable. Thus, using linear models may result in limited predictive performance.

Tree-based methods are data-driven tools based on sequential procedures that recursively partition the data. These methods offer straightforward approaches to grasp and summarise the key features of the data and leverage tree graphs to offer visual depictions of the rules governing a specific dataset \cite{noauthor_15_nodate}. Hence, we select five tree-based ML models, namely HistGradientBoosting Classifier, RandomForest Classifier, XGBClassifier, LGBMClassifier, and CatBoostClassifier, that can manage the complexity and sparsity of the dataset, providing a balance between interpretability, efficiency, and predictive performance through various mechanisms, e.g., ensembles, histogram-based, leaf-wise, or symmetric tree growth algorithms.


DL research in the context of classification tasks in tabular data has witnessed significant advancements in recent years, yielding a few state-of-the-art models like TabTransformer \cite{huang_tabtransformer_2020}. Multiple experiments demonstrated that TabTransformer surpasses \gls{MLP} and recent deep networks designed for tabular data on some datasets, however tree-based ensemble models like GBDT (Gradient Boosting Decision Trees) still perform better on the majority of tested datasets and come with shorter training time \cite{huang_tabtransformer_2020,borisov_deep_2024,gardner_benchmarking_nodate,mcelfresh_when_nodate,gorishniy_revisiting_2021}.


TabTransformer introduces a novel self-attention deep tabular data modeling architecture for supervised and semi-supervised learning. It enhances prediction accuracy by transforming categorical feature embeddings into robust contextual embeddings \cite{huang_tabtransformer_2020}. The presented architecture is highly robust against missing and noisy data features, making it a promising model for predicting an appropriate level of care using questionnaire-based, highly sparse, and highly dimensional triage data. We select TabTransformer to benchmark against the tree-based models.

\subsection{Experiments design}

The study included the following four series of experiments. We conducted the first three series on datasets of completed triage interviews to select, train, and tune ML models. We performed the final fourth series of experiments to evaluate the models' inference performance on unfinished interviews. 
\begin{enumerate}
    \item Choosing the dataset type (NaN or non-NaN) and ML models for subsequent evaluation experiments.
    \item Training the models and evaluating their performance with default hyperparameters on datasets with completed interviews, selecting models for further hyperparameter optimization.
    \item Evaluating the performance of tuned models on datasets of completed interviews.
    \item Assessing the inference performance (prediction accuracy) of the models on datasets of unfinished interviews with various degrees of completeness.
\end{enumerate}
In each experiment, we used an 80/20 train-test split, stratified by the target feature, and ran five times with a random split to ensure consistent results. We use the balanced accuracy score, an evaluation metric for binary and multiclass classification problems that help deal with imbalanced target class distribution. 

We perform the first series of experiments to choose the more effective representation of missing or undefined values in the triage interview dataset from two possible options: either NaN (Not a Number) or non-NaN (0 for numeric features or False for Boolean) to represent missing (empty) answers and, finally, to select ML models for the subsequent experiments. This series of experiments included evaluating ML models on an experimental non-NaN dataset and an experimental NaN dataset and evaluating the DL model, namely TabTransformer, on an experimental NaN dataset.

The second series of experiments aims to evaluate the models' performance with default parameters and choose models for hyperparameter optimization. In contrast, the third series aims to evaluate the performance of tuned models.

The final series 
includes experiments on datasets with incomplete interviews with the best-performing ML models. 
To imitate a real-world scenario of an unfinished triage interview, we create incomplete interviews by progressively removing answers, starting from the last responses of the complete interviews from 100\% to 40\%, decreasing by 20\% each time. This results in four datasets with various degrees of interview completeness: 100\%, 80\%, 60\%, and 40\%. 

\section{Results}
\label{sec:results}
\subsection{Performance of ML Models}

The first experimental stage included running ML models on the experimental dataset with approximately 15,000 interviews and over 2,000 features. Tables~\ref{tab:nonNanResults} and~\ref{tab:nanSmallResults} represent the average results of running models on non-NaN and NaN datasets retrospectively.

\begin{table}
  \begin{center}
    \caption{Accuracy and training/inference time of studies models on the experimental non-NaN dataset of completed interviews.}
    \label{tab:nonNanResults}
    \resizebox{\columnwidth}{!}{%
    \begin{tabular}{c|c|c|c|c}
      \textbf{Model} & \makecell{\B Avg training \\ \B accuracy, \%} & \makecell{ \B Avg test \\ \B accuracy, \%} & \makecell{\B Avg training\\ \B time, min} & \makecell{\B Avg inference\\ \B time, min} \\
      \hline
      \makecell{HistGradientBoosting\\Classifier} & 76,6 & 68,2 & 0,57 & 0,005  \\
      \hline
      \makecell{RandomForest\\Classifier} & \B 99,2 & 78,6 & 0,03 & 0,003 \\
      \hline
      XGBClassifier & 83,2 & 74,2 & 0,72 & 0,003 \\
      \hline
      LGBMClassifier & 79,3 & 70,8 & 0,05 & 0,003 \\
      \hline
      CatBoostClassifier & 73,2 & 67,9 & 0,54 & \B 0,001 \\
    \end{tabular}
    }
  \end{center}
\end{table}

\begin{table}
  \begin{center}
    \caption{Accuracy and training/inference time of studies models on the experimental NaN dataset of completed interviews.}
    \label{tab:nanSmallResults}
    \resizebox{\columnwidth}{!}{%
    \begin{tabular}{c|c|c|c|c}
      \textbf{Model} & \makecell{\B Avg train \\ \B accuracy, \%} & \makecell{ \B Avg test \\ \B accuracy, \%} & \makecell{\B Avg training\\ \B time, min} & \makecell{\B Avg inference\\ \B time, min} \\
      \hline
      \makecell{HistGradientBoosting\\Classifier} & 86,0 & 76,6 & 0,65 & 0,003  \\
      \hline
      \makecell{RandomForest\\Classifier} & \B 95,9 & 82,0 & \B 0,025 & 0,002 \\
      \hline
      XGBClassifier & 88,4 & 80,4 & 0,03 & 0,002 \\
      \hline
      LGBMClassifier & 91,8 & \B 82,2 & 0,07 & 0,003 \\
      \hline
      CatBoostClassifier & 78,7 & 73,1 & 0,12 & \B 0,001 \\
      \hdashline[2pt/2pt]
      \hdashline[2pt/2pt]
      TabTransformer & \B 99,98 & \B 99,97 & 700,6 & 0,56 \\
    \end{tabular}
    }
  \end{center}
\end{table}

According to these experiments, using a NaN dataset is more beneficial since it provides higher accuracy and faster training time. Thus, the subsequent experiments used a training dataset that contained NaN values. Table~\ref{tab:nanDefaultResults} shows the results of running ML models with default parameters on the full dataset.

\begin{table}
  \begin{center}
    \caption{Accuracy and training/inference time of studies models with default parameters on the completed interview dataset.}
    \label{tab:nanDefaultResults}
    \resizebox{\columnwidth}{!}{%
    \begin{tabular}{c|c|c|c|c}
       \textbf{Model} & \makecell{\B Avg train \\ \B accuracy, \%} & \makecell{ \B Avg test \\ \B accuracy, \%} & \makecell{\B Avg training\\ \B time, min} & \makecell{\B Avg inference\\ \B time, min} \\
      \hline
      \makecell{HistGradientBoosting\\Classifier} & 90,6 & 89,9 & 5,19 & 0,04  \\
      \hline
      \makecell{RandomForest\\Classifier} & \B 94,7 & 89,3 & 1,29 & 0,05 \\
      \hline
      XGBClassifier & 87,4 & 86,5 & \B 0,22 & 0,03 \\
      \hline
      LGBMClassifier & 92,0 & \B 91,3 & 0,49 & 0,09 \\
      \hline
      CatBoostClassifier & 89,1 & 88,5 & 0,85 & \B 0,005 \\
      \hdashline[2pt/2pt]
      \hdashline[2pt/2pt]
      TabTransformer (7\%) & \B 76,4  & \B 74,4 & 1823 & 0,49 \\
    \end{tabular}
    }
  \end{center}
\end{table}

All five machine learning algorithms demonstrated comparable performance when applied to the complete training dataset. Consequently, we decided to conduct further hyperparameter optimization for each algorithm. Table~\ref{tab:tunedMlResults} presents the performance of these algorithms after tuning the parameters and training the models on the entire training dataset.

\begin{table}
  \begin{center}
    \caption{Accuracy and training/inference time of studies models with tuned parameters on the completed interview dataset.}
    \label{tab:tunedMlResults}
    \resizebox{\columnwidth}{!}{%
    \begin{tabular}{c|c|c|c|c}
      \textbf{Model} & \makecell{\B Avg train \\ \B accuracy, \%} & \makecell{ \B Avg test \\ \B accuracy, \%} & \makecell{\B Avg training\\ \B time, min} & \makecell{\B Avg inference\\ \B time, min} \\
      \hline
      \makecell{HistGradientBoosting\\Classifier} & 95,5 & 94,2 & 7,13 & 0,048  \\ 
      \hline
      \makecell{RandomForest\\Classifier} & 86,6 & 82,8 & 2,68 & 0,054 \\
      \hline
      XGBClassifier & 95,3 & 93,6 & \B 0,36 & 0,030 \\
      \hline
      LGBMClassifier & \B 99,5 & \B 97,3 & 1,19 & 0,151 \\
      \hline
      CatBoostClassifier & 98,7 & 96,5 & 43,2 & \B 0,007 \\
    \end{tabular}
    }
  \end{center}
\end{table}

Due to the limited computational resources, running \gls{DL} models on a full dataset was not feasible. Instead, we ran only on the experimental dataset and a subset (7\%) of the full dataset.

Tables~\ref{tab:nanSmallResults} and~\ref{tab:nanDefaultResults} present the results of running \gls{DL} models with default parameters.



\subsection{Comparison and Analysis}

Table~\ref{tab:tuningMlComparison} shows that all tested decision trees performed relatively well, with over 80\% accuracy. However, the RandomForestClassifier surprisingly performed worse with the tuned parameters. The possible cause is the lack of computational resources that limit the sample variability of tuned parameters. Thus, the RandomForestClassifier used default parameters for further evaluation.

\begin{table}
  \begin{center}
    \caption{Accuracy of studied models on the completed interview dataset before and after tuning.}
    \label{tab:tuningMlComparison}
    \resizebox{\columnwidth}{!}{%
    \begin{tabular}{c|c|c}
      \textbf{Model} & \textbf{Avg test accuracy before, \%} & \textbf{Avg test accuracy after, \%} \\
      \hline
      \makecell{HistGradientBoosting\\Classifier} & 89,9 & 94,2 \\
      \hline
      \makecell{RandomForest\\Classifier} & 89,3 & 82,8 \\
      \hline
      XGBClassifier & 86,5 & 93,6 \\
      \hline
      LGBMClassifier & 91,3 & 97,3 \\
      \hline
      CatBoostClassifier & 88,5 & 96,5 \\
    \end{tabular}
    }
  \end{center}
\end{table}

\subsection{Correlation between completeness and prediction accuracy}

The study evaluated tuned models' performance on test datasets with varying degrees of completeness and the corresponding average accuracy. We use the best-performing parameters from the previous step. We employed the strictest approach of defining the accuracy by using the original outcome as the only correct option. In practice, there can be multiple correct outcomes as the degree of incompleteness and variability of missing answers grow. However, the study lacked a reliable source to identify these alternative correct options, so we relied solely on the original outcomes for accuracy assessment.

\begin{figure}
  \begin{center}
    \includegraphics[width=0.42\textwidth]{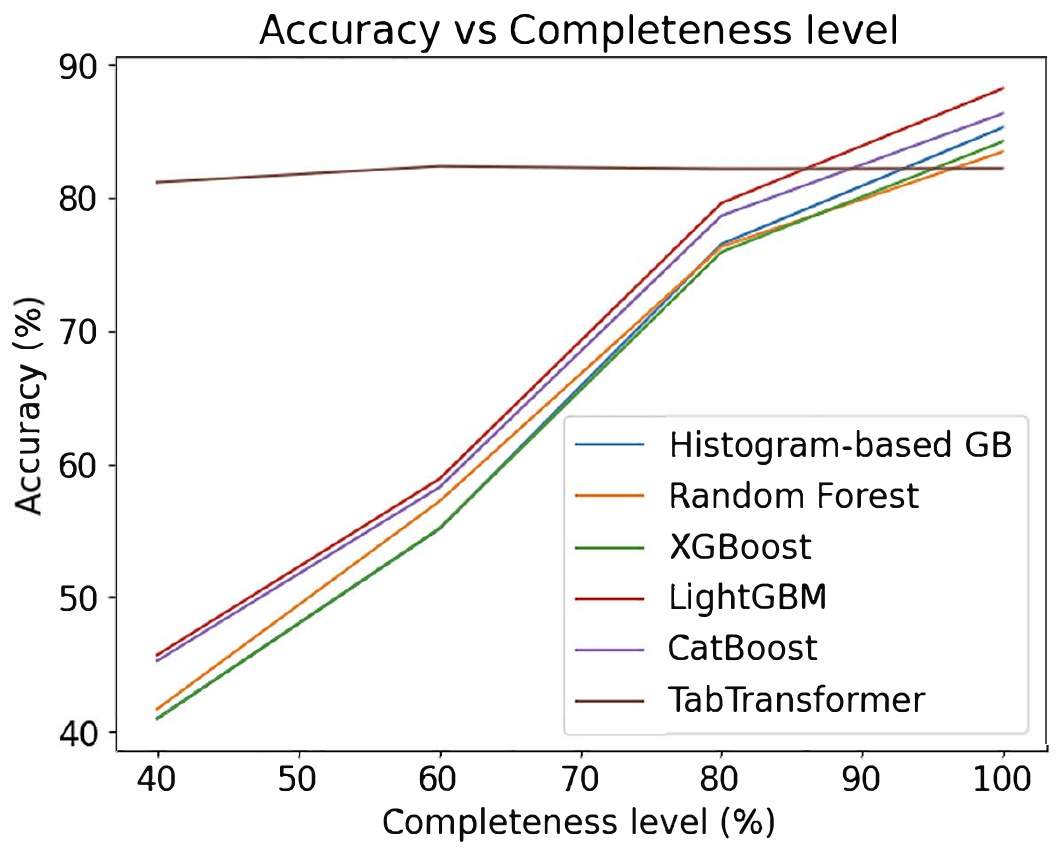}
  \end{center}
  \caption{Correlation between completeness level and prediction accuracy}
  \label{fig:finalEval}
\end{figure}

Table~\ref{tab:finalEval} shows the accuracy versus the completeness level. Figure~\ref{fig:finalEval} also visually depicts how the accuracy degrades with decreasing levels of completeness. All evaluated decision-tree models had similar performance rates. However, LGBMClassifier and CatBoostClassifier performed best on all test datasets, having noticeably higher accuracy on a test dataset with 40\% completeness. Interestingly, TabTransformer maintains accuracy at the same level for all the completeness degrees, which TabTransformer's self-attention mechanism might explain, handling the incomplete data remarkably.

\begin{table}
  \begin{center}
    \caption{Accuracy of studies models on the dataset of incompleted interviews with various degrees of completeness.}
    \label{tab:finalEval}
    \begin{tabular}{c|c|c|c|c}
      \multirow{2}{*}{\textbf{Model}} & \multicolumn{4}{c}{\textbf{Completeness Level}} \\
      \cline{2-5}
      & \textbf{100\%} & \textbf{80\%} & \textbf{60\%} & \textbf{40\%} \\
      \hline
      \makecell{HistGradientBoosting\\Classifier} & 85,3 & 76,5 & 55,2 & 41,0 \\
      \hline
      \makecell{RandomForest\\Classifier} & 83,5 & 76,3 & 57,3 & 41,7 \\
      \hline
      XGBClassifier & 84,3 & 76,0 & 55,2 & 41,0 \\
      \hline
      LGBMClassifier & \B 88,2 & \B 79,6 & \B 58,9 & \B 45,7 \\
      \hline
      CatBoostClassifier & \B 86,4 & \B 78,7 & \B 58,3 & \B 45,3 \\
      \hdashline[2pt/2pt]
      \hdashline[2pt/2pt]
      TabTransformer & 82.2  & \B 82.2  & \B 82.3 & \B 81.2 \\
    \end{tabular}
  \end{center}
\end{table}

\section{Discussion}
\label{sec:discussions}
\subsection{Interpretation of Results}

As a result of the study, we shortlisted a few decision-tree models as the most efficient. We evaluated the performance of each model on different data subsets:

\begin{enumerate}
    \item Test subset (from train/test split) to evaluate a model after the training step.
    \item Evaluation subset to define the baseline for unfinished interviews.
\end{enumerate}

According to the results shown in Table~\ref{tab:trainVsEval}, the test accuracy of the decision trees at the training step is noticeably higher than at the evaluation step. This discrepancy may result from data variability in subsets. Since the only condition for splitting the full dataset into subsets was stratification by the interview outcome, different subsets may have captured interviews from different questionnaires, age groups, or demographics, introducing sample bias. However, the model performance is over 80\% even at the evaluation step, indicating that the models could still generalize effectively.

\begin{table}
  \begin{center}
  
    \caption{Accuracy of studies models at the training vs. evaluation step.}
    \begin{tabular}{c|c|c}
      \textbf{Model} & \textbf{Training step} & \textbf{Evaluation step} \\
      \hline
      HistGradientBoostingClassifier & 94,2 & 85,3 \\
      \hline
      RandomForestClassifier & 82,8 & 83,5 \\
      \hline
      XGBClassifier & 93,6 & 84,3 \\
      \hline
      LGBMClassifier & \B 97,3 & \B 88,2 \\
      \hline
      CatBoostClassifier & \B 96,5 & \B 86,4 \\
      \hdashline[2pt/2pt]
      \hdashline[2pt/2pt]
      TabTransformer & 74.4  & 82.2 \\
    \end{tabular}
    \label{tab:trainVsEval}
  \end{center}
\end{table}

Even though each model's performance is similar, LGBMClassifier and CatBoostClassifier consistently outperform other models. Both these algorithms are powerful gradient-boosting classifiers commonly used in ML. LGBMClassifier provides faster training time and accuracy, although categorical and text data must be encoded numerically before training. 
Meanwhile, CatBoostClassifier automatically handles categorical features and missing values without preprocessing, providing faster inference time. Thus, both models employ beneficial properties for dealing with the given dataset, and the study recommends using these classifiers for potential integration with the existing system.

Another noticeable outcome of this study is TabTransformer's performance. Being trained only on a small dataset, TabTransformer showed spectacular performance — over 80\% accuracy on all unfinished interviews. Such exceptional performance is likely due to the self-attention mechanism employed by this model, making it highly robust against missing and noisy features in the given interview dataset. However, the cost for such good performance is training time — over 30 hours on local hardware. The study recommends acquiring more powerful hardware, like cloud-based GPU, and investigating the model performance on a full dataset.

It is important to note that interviews are dynamically constructed with varying lengths based on the responses provided. Consequently, it is impossible to determine the number of unanswered questions accurately. The ideal completeness threshold should be established by analyzing historical data and considering the safety requirements of healthcare units.

\subsection{Challenges and Limitations}


Sparse, questionnaire-based data presents specific challenges for ML model development, particularly in preprocessing. Missing values can bias results and reduce performance, but they might also contain useful information. Only a few models can handle missing data effectively. In this study, we first determined the best approach to handle missing values: whether to impute them or retain NaNs. As shown in Tables~\ref{tab:nonNanResults} and~\ref{tab:nanSmallResults}, using a non-NaN dataset increased training time without improving performance.



We conducted the study on a MacBook 16" with a built-in GPU and 32 GB RAM, constrained by privacy regulations against using public cloud services for training on real-world digital triage data. Although the M1 chip offers good performance and efficiency, it is insufficient for large datasets or complex algorithms requiring intensive computation. Therefore, the study was limited in its exploration of model performance on a full dataset and incomplete interview instances.


\section{Conclusion and Future Work}
\label{sec:conclusion}


The study explored the challenges and considerations involved in developing an ML system for defining the outcome of triage interviews based on sparse questionnaire data. The study revealed that sparse, questionnaire-based data presents unique challenges throughout the ML pipeline, from data preprocessing to model selection and integration, namely, limitations in handling missing values, selecting appropriate ML models, and scaling up to handle large datasets or complex ML algorithms. Despite these challenges, the study identified several key findings and recommendations:

\begin{enumerate}
    \item We identified the dataset with NaN values as a more efficient approach for handling missing values than imputation with non-empty values. Empirical evidence also demonstrated better performance of models that support NaN values, such as decision-tree models.
    \item Decision-tree models emerged as the most efficient approach for the study, given their ability to handle large, sparse datasets containing empty values.
    \item The LGBMClassifier and CatBoostClassifier consistently outperformed other models across different data subsets. TabTransformer was also identified as a promising approach to learning and predicting based on sparse and complex data structures, albeit with significantly longer training times, indicating the potential benefits of cloud-based GPU infrastructure for training complex models on larger datasets.
    \item The study also identified a linear correlation between interview completeness degree and the predictive power of studied decision-tree ML models, such as LGBMClassifier and CatBoostClassifier.
\end{enumerate}

In conclusion, the study contributes valuable insights into developing ML systems in medical triage to predict the interview outcome by analyzing sparse, questionnaire-based data. While conducted on limited local hardware, the findings underscore the potential benefits of leveraging a more powerful infrastructure for training complex models on large datasets. Due to the problem's breadth, 
future research areas include:

\begin{enumerate}
    \item Optimizing storage and processing of sparse data to improve model training/inference time and performance, reducing system resource consumption.
    \item Exploring advanced model architectures like TabTransformer that require more powerful computational resources but offer higher predictive performance crucial for error-sensitive domains like medicine.
    \item Investigating techniques to generate interpretable explanations for model predictions and identifying key features to increase trust and acceptance of ML systems among healthcare practitioners and patients.
    \item Treating interview questions and answers as a time-series classification problem, given their ordered nature in designed questionnaires.
\end{enumerate}

\bibliographystyle{IEEEtran}
\bibliography{bibliography}

\end{document}